\documentclass[letterpaper, 10 pt, conference]{ieeeconf}  

\IEEEoverridecommandlockouts                            

\let\labelindent\relax

\usepackage{enumitem}

\usepackage{graphics} 
\usepackage{epsfig} 
\usepackage{mathptmx} 
\usepackage{times} 
\usepackage{amsmath} 
\usepackage{amssymb}  
\usepackage{outlines}
\usepackage{xcolor}
\usepackage[ruled,vlined,linesnumbered]{algorithm2e}
\usepackage{gensymb}
\usepackage{caption}
\usepackage{float}
\usepackage{enumitem}
\usepackage{subcaption}
\usepackage[colorlinks=true,linkcolor=black,anchorcolor=black,citecolor=black,filecolor=black,menucolor=black,runcolor=black,urlcolor=black]{hyperref}
\usepackage{siunitx}
\usepackage{newtxmath}  
\usepackage{mathtools}
\usepackage{cite}

\title{\LARGE \bf
Neural Elevation Models for Terrain Mapping and Path Planning
}

\author{Adam Dai$^{1}$, Shubh Gupta$^{2}$ and Grace Gao$^{2}$
\thanks{$^{1}$Department of Electrical Engineering,
        Stanford, CA 94305, USA,
        addai@stanford.edu}%
\thanks{$^{2}$Department of Aeronautics and Astronautics,
        Stanford, CA 94305, USA
        \{shubhgup, gracegao\}@stanford.edu}%
}

\newcommand{\R}{\ensuremath{\mathbb{R}}}

\newcommand{\p}{\mathcal{P}}
\newcommand{\height}{\mathcal{H}}

\newcommand{\mbf}[1]{{\mathbf{#1}}}

\newcommand{\norm}[1]{\left\Vert#1\right\Vert}

\begin{document}

\maketitle
\thispagestyle{empty}
\pagestyle{empty}

\begin{abstract}


This work introduces Neural Elevations Models (NEMos), which adapt Neural Radiance Fields to a 2.5D continuous and differentiable terrain model.
In contrast to traditional terrain representations such as digital elevation models, NEMos can be readily generated from imagery, a low-cost data source, and provide a lightweight representation of terrain through an implicit continuous and differentiable height field.
We propose a novel method for jointly training a height field and radiance field within a NeRF framework, leveraging quantile regression. 
Additionally, we introduce a path planning algorithm that performs gradient-based optimization of a continuous cost function for minimizing distance, slope changes, and control effort, enabled by differentiability of the height field.
We perform experiments on simulated and real-world terrain imagery, demonstrating NEMos ability to generate high-quality reconstructions and produce smoother paths compared to discrete path planning methods.
Future work will explore the incorporation of features and semantics into the height field, creating a generalized terrain model. 

\end{abstract}

\section{Introduction}

\begin{figure}
    \centering
    \includegraphics[width=0.43\textwidth]{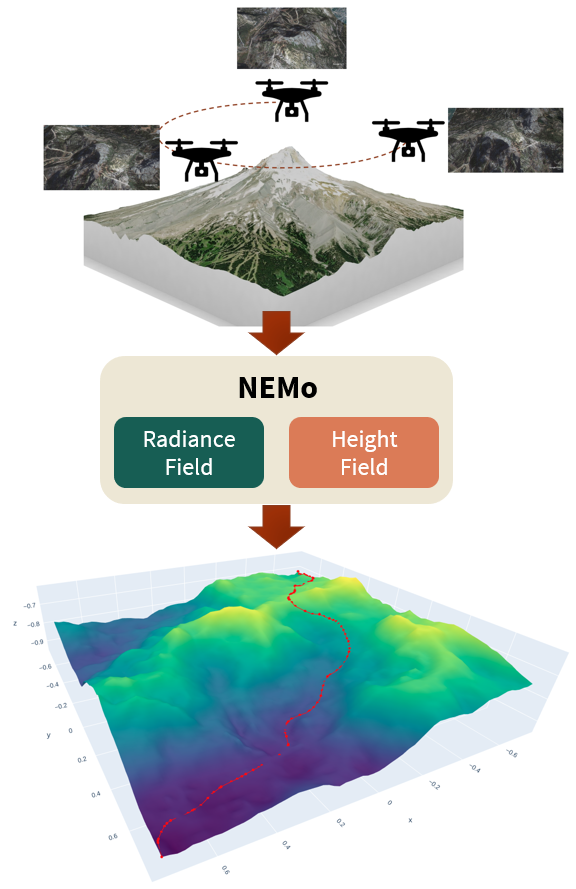}
    \caption{Neural Elevation Model (NEMo) framework. Aerial imagery of terrain is captured and used to simultaneously train a radiance field and height field of the scene. The height field is then used as a continuous and differentiable map representation for gradient-based path planning.}
    \label{fig:overview}
\end{figure}



Autonomous ground robots operating in challenging terrain play a vital role in various applications, such as search and rescue operations in disaster zones, infrastructure inspection in remote areas, and planetary exploration. These environments are highly diverse and complex, with uneven surfaces, loose materials, and difficult-to-model terramechanics posing significant risks to the robot's functionality and mission success if their intricacies are not factored into navigation. Therefore, there is a need for navigation methods that can both effectively model these environments and leverage those models for safe and efficient navigation.

Traditionally, terrain details have been captured through expensive and time-consuming high-fidelity surveys using 3D sensors like LiDAR and Radar. Ranging data is then used to generate a Digital Elevation Model (DEM) of the terrain. DEMs represent elevation data using either a discrete grid of squares (raster) or a network of interconnected triangles (mesh). 
Path planning typically consists of generating a costmap from the DEM based on slope and roughness and using global search algorithms such as A* or Field D*~\cite{ono2016data, carsten2007global}.
While high-fidelity DEMs provide detailed representation of terrain, their reliance on expensive sensors makes them prohibitive to use at scale and on changing terrain.

To address the limitations of expensive 3D sensors, methods like multi-view stereo (MVS) \cite{schoenberger2016sfm} and photogrammetry rely on inexpensive cameras to reconstruct 3D environments. These techniques leverage multiple images of the terrain captured from different viewpoints. 
However, these methods rely on fixed, discrete representations (e.g., point clouds or meshes), limiting their resolution and potentially leading to artifacts when capturing intricate details. Additionally, they are highly sensitive to image quality (e.g., lighting, occlusion, texture) and can be computationally demanding, especially for large-scale scenes.  

Recently, Neural Radiance Fields (NeRFs) \cite{mildenhall2021nerf} have emerged as powerful tools for environment representation using camera images. NeRFs can be directly generated from posed images and have demonstrated remarkable performance in capturing visual detail in challenging conditions~\cite{gao2022nerf}. 
Compared to DEMs, NeRFs offer several advantages: faster generation, denser reconstruction, and a compact yet continuous and differentiable representation. However, their usability for terrain navigation suffers from two key limitations:
\begin{itemize}
    \item \textbf{High Processing Overhead:} Unlike DEMs, which provide interpretable height information associated with a ground coordinate, NeRFs capture a full scene with implicit density and color. This richer representation, while powerful, requires additional processing steps to extract the relevant terrain data for informing navigation.
    \item \textbf{Focus on Complex Scenes:} NeRFs were initially designed to represent a variety of complex 3D scenes, including visual properties of buildings and several non-terrain elements~\cite{mari2022sat, xiangli2022bungeenerf}. This can introduce unnecessary complexity for ground robots solely focused on navigating the terrain.   
\end{itemize}
These limitations highlight the potential for combining both approaches. By leveraging the strengths of NeRFs in capturing complex terrain detail and the suitability of DEMs for path planning, we can create more robust and versatile navigation solutions.


In this work, we introduce Neural Elevation Models (NEMos), which integrate a Neural Radiance Field alongside a height field.
We propose a novel approach for jointly training the NeRF and the height field, in which height is learned from the NeRF through quantile regression, and, in turn, the height network supervises the NeRF to eliminate spurious density above the surface.
We then demonstrate a method for path planning that leverages the continuous and differentiable nature of the height network to achieve smoother paths than those obtained via discrete planning over equivalent DEMs.
The field of Neural Elevation Models is still nascent, so we demonstrate initial results in this paper and discuss directions for future research and exploration.

\section{Terrain Mapping}


Our objective is to construct a map of terrain from aerial images $\{I_1,\dots,I_N\}$ with associated camera poses $\{T_1,\dots,T_N\}$.
We assume that the camera poses are provided in an East-North-Up (ENU) frame. This ensures that the scene's ground plane aligns with the $XY$ plane in the NeRF coordinate system, which is crucial for accurate elevation estimation during terrain reconstruction.

\subsection{Neural Elevation Model (NEMo)}

NEMo offers a novel approach to continuous terrain elevation representation, allowing direct generation from 2D camera images without additional depth data. 

NEMo simultaneously trains two models: 
\begin{itemize}
    \item A \textit{NeRF model} captures the overall 3D scene with density and color information. We use $\rho : \R^3 \rightarrow \R^+$ to refer to the NeRF density.
    \item A \textit{height field}, parametrized using a neural network, specifically focuses on predicting terrain elevation. The height field is denoted as $\height : \R^2 \rightarrow \R$, which maps any ground coordinate $(x,y) \in \R^2$ to a corresponding height value $z \in \R$ in the scene.
\end{itemize}
After the training, the NeRF component is discarded and only the height field is retained to maintain a compact representation.




\subsection{Training}

\subsubsection{NeRF Training}

NEMo employs a standard training procedure for the NeRF component, minimizing a volumetric rendering loss function \cite{tancik2023nerfstudio}. Additionally, the estimated height from the height network is used to mask out regions that exceed the predicted elevation. This effectively removes artifacts or ``floaters" above the actual terrain, allowing the NeRF to focus on the surface for accurate scene reconstruction. We observed that this masking process during the NeRF training leads to a cleaner NeRF representation with fewer artifacts.
\subsubsection{Height Network Training}

\begin{figure}
    \centering
    \includegraphics[width=0.49\textwidth]{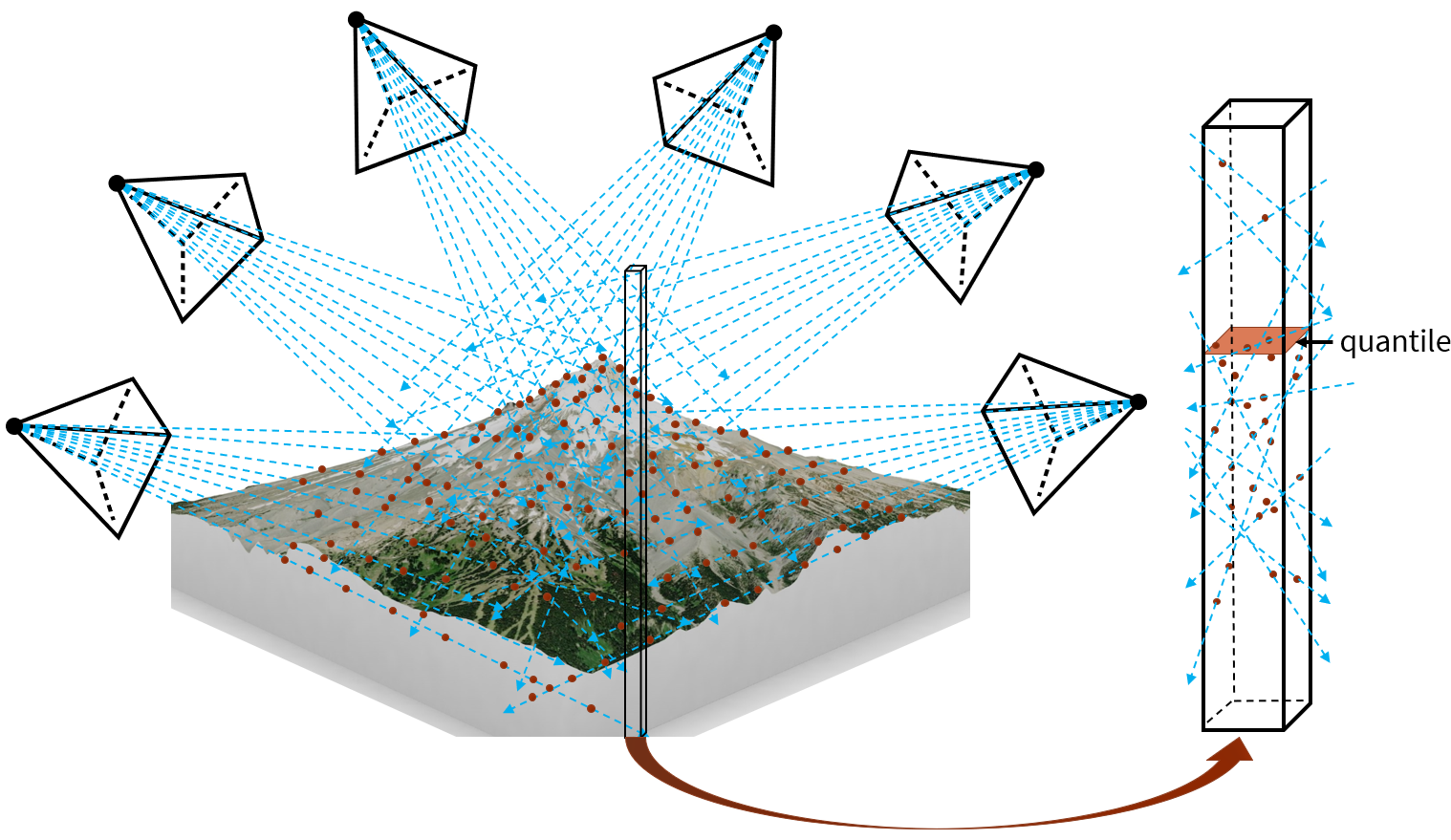}
    \caption{Height field training via quantile regression. Training images cast rays (depicted in light blue) into the scene, along which points are sampled (dark red). By minimizing the weighted quantile loss over the multitude of sampled points, we effectively regress the desired quantile of vertical NeRF density as height.}
    \label{fig:quantile_reg}
\end{figure}

The height network leverages quantile regression~\cite{koenker2001quantile} to learn the height information from images alongside the NeRF density distribution. During training, for each ray cast onto the scene, we sample $n$ 3D points $\{(x_1,y_1,z_1),\dots,(x_n,y_n,z_n)\}$ along its path. We then compute a quantile loss $\mathcal{L}_q$ based on the error $e_i = z_i - \hat{z}_i$ between height $z_i$ of the $i$-th point and predicted height $\hat{z}_i$ from the network. To prioritize the information from occupied space, we weight the loss using a factor $w_i = T_i \alpha_i$, where $T_i$ is transmittance and $\alpha_i$ is opacity, which are computed along the ray based on the learned density $\rho$. The total loss $\mathcal{L}$ as a function of the predicted heights $\hat{\mbf{z}}$ for training the height network is obtained as follows:
\begin{align}
    \mathcal{L}(\hat{\mbf{z}}) &= \sum_{i=1}^n w_i \mathcal{L}_q(z_i, \hat{z}_i) \\
    \mathcal{L}_q(z_i, \hat{z}_i) &=
    \left\{
    	\begin{array}{ll}
    		-(1-q) \cdot e_i  & \mbox{if } e_i < 0 \\
    		q \cdot e_i       & \mbox{if } e_i \geq 0
    	\end{array}
    \right.
\end{align}
Intuitively, when the error is negative and $\hat{z}_i$ is greater than $z_i$, the loss pulls the estimate $\hat{z}_i$ down with strength proportional to $(1-q)$. 
Conversely, when the error is positive and $\hat{z}_i$ is less than $z_i$, the loss pulls the estimate up with strength proportional to $q$.
This loss, also known as the ``pinball loss," has been shown to converge to the quantile $q$~\cite{yang2016posterior}.
The quantile regression process is illustrated in Fig. \ref{fig:quantile_reg}.

\section{Path Planning}

Now, given trained height field $\height: (x,y) \rightarrow z$, a start location $(x_0, y_0)$, and goal location $(x_f, y_f)$, our objective is to plan a path $\p$ from start to goal which minimizes (1) distance traveled, (2) slope along the path, and (3) necessary control effort.
The implicit nature of $\height$ allows us to formulate a cost function and optimize for this path in continous space. 
Our planning approach is most similar to that of \cite{adamkiewicz2022vision}, in which a path is initialized with A*, then optimized with gradient descent for control cost and collision avoidance---however, in our case, we penalize path slope from our height field as opposed to collision avoidance from the NeRF density field. 

\subsection{Initialization}
In practice the cost manifold possesses many local minima, and thus we first initialize with A* over a discretized version of $\height$ to obtain an initial solution in the neighborhood of the global optimum.
We use change in height between cells for cost, and thus A* attempts to optimize for objectives (1) and (2).
Note that the worst-case runtime of A* scales roughly exponentially with respect to grid size, and thus it is prohibitively expensive to run on larger grids. 

\subsection{Continuous Path Optimization} 
Next, we leverage the fact that $\height$ allows us to evaluate the height and gradients (through automatic differentiation) of an arbitrary point in continuous space to formulate the path optimization over a continuous path.
For curve $\p : [0, T] \rightarrow \R^2$ which represents a 2D path,
we parameterize $\p$ with a series of differentially flat output waypoints $W=\{\sigma_1,\dots,\sigma_N\}$ under dynamics $\dot{x} = f(x,u)$.
The dynamics can be used to integrate between waypoints and evaluate $\p$ at any $\tau \in [0, T]$, thus providing a continuous representation of the path, $\p(\tau)$.

Now, in terms of $\p(\tau)$, we formulate three cost terms $J_1, J_2, J_3$ for minimizing the three desired path objectives mentioned above:
\begin{enumerate}[label=(\arabic*)]
    \item \textit{Path distance}: This is computed as arc length by integrating path speed $\norm{\frac{\partial \p}{\partial t}}$ over the path. Thus, 
    \begin{equation}
        J_1(\tau) = \norm{\frac{\partial \p}{\partial t}\bigg|_{\tau}}.
    \end{equation}
    \item \textit{Slope along path}: The slope of the terrain along the path at $\p(\tau)$ is the directional derivative of $\height$ in the path direction $\frac{\partial \p}{\partial t}$.
    Thus,
    \begin{equation}
        J_2(\tau) = \frac{\partial \p}{\partial t}\bigg|_{\tau} \cdot \nabla \height(\p(\tau)).
    \end{equation}
    \vspace{1cm}
    \item \textit{Control effort}: 
    The control $u(\tau)$ at $\tau$ is determined from the differential flatness mapping \cite{levine2011necessary}. We then write
    \begin{equation}
        J_3(\tau) = u(\tau)^T R u(\tau),
    \end{equation}
    where $R$ is a weighting matrix.
\end{enumerate}
The total path cost is expressed as the integral over the path of the weighted sum of these cost terms:
\begin{equation}
\begin{aligned}
    J_{total} = \int_0^T \beta_1 J_1(\tau) + \beta_2 J_2(\tau) + J_3(\tau) \ dt
\end{aligned}
\end{equation}
where $\beta_1$ and $\beta_2$ are weighting scalars.

This cost function is optimized over the waypoints $W$ which parameterize $\p$ to obtain optimal waypoints $W^*$ that parameterize the refined path $\p^*$
\begin{equation}
\begin{aligned}
    W^* = \min_{W} \ J_{total}(W).
\end{aligned}
\end{equation}

In practice, to perform optimization we still need to sample points on the path, but the continuous formulation allows us to do so at a much higher resolution for cost evaluation, while still optimizing only over the sparse waypoints $W$.
For future work, we plan to include additional objectives to optimize for traversability based on semantics and features extracted from the NeRF.



\section{Results}

For our experiments\footnote{Our code is available at: \url{https://github.com/Stanford-NavLab/nerfstudio/tree/adam/terrain} (NEMo model and training) and \url{https://github.com/adamdai/neural_elevation_models} (analysis and path planning).}, we consider two different terrain scenes, one from simulated imagery and one from real-world imagery: 
\begin{enumerate}
    \item ``\textbf{KT-22}": Imagery and camera poses collected from Google Earth Studio~\cite{GES} of the KT-22 peak in Olympic Valley, CA.
    \item ``\textbf{Red Rocks}": Drone imagery collected by Falcon fixed-wing UAV over Red Rocks, CO \cite{dronemapper}. COLMAP \cite{schoenberger2016sfm} is used to estimate camera poses.
\end{enumerate}
For each scene, we train a NEMo and analyze the quality of the NeRF and height field. 
We then use the height field to plan paths over the terrain and evaluate them with metrics for distance, average slope, and smoothness.

\subsection{NEMo Training}

\begin{figure}
    \centering
    \includegraphics[width=0.5\textwidth]{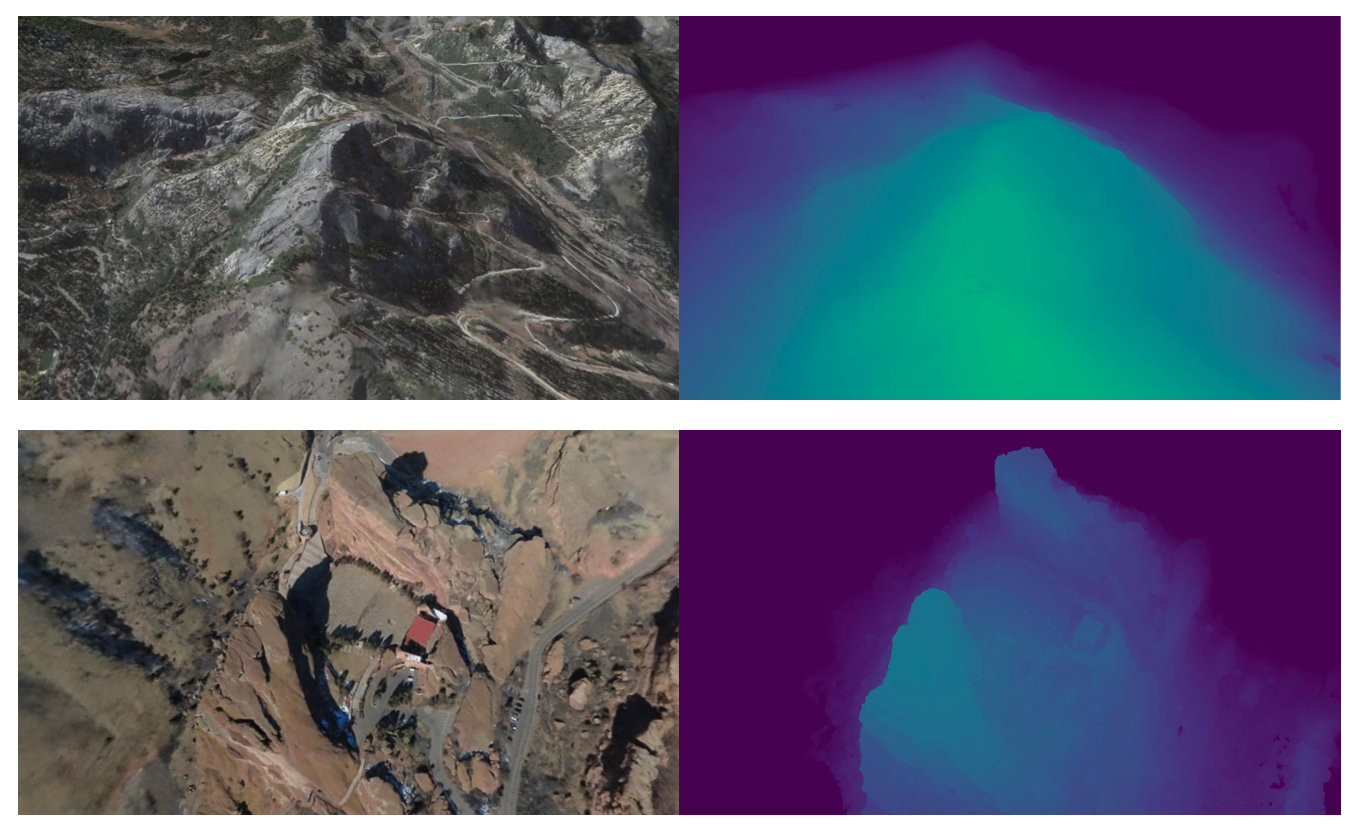}
    \caption{RGB (left) and depth (right) renders from trained NEMos of the scenes used in our experiments. KT-22 is shown on the top row, and Red Rocks on the bottow row.}
    \label{fig:renders}
\end{figure}

The NeRF component of NEMo is based on the Nerfacto model from Nerfstudio \cite{tancik2023nerfstudio}.
The height network is implemented as a hashgrid encoding \cite{muller2022instant} followed by a single-layer MLP with ReLU activation.
Fig. \ref{fig:renders} shows RGB and depth renders from the trained NeRF component of NEMos for each scene, demonstrating detailed reconstruction with underlying geometry.
Anecdotally, we also observe that NEMo NeRFs possess less haze above the scene than those trained with standard Nerfacto, due to the height field masking out spurious density above the ground in free space.


We note that our approach for elevation estimation and representation differs from training a standard NeRF then rendering its top down density as a height map, in that:
(1) a render must sample points and thus is an explicit and discrete representation whereas the height network is an implicit function, and (2) the NEMo jointly trains the NeRF and height network, allowing the NeRF to benefit from height supervision.


\subsection{Path Planning}

For path optimization, we use a Dubin's car dynamics model for differential flatness, and employ the Adam \cite{kingma2014adam} optimizer with learning rate $\SI{1e-3}{}$.
\begin{figure}[!ht]
     \centering
     \begin{subfigure}[b]{0.47\textwidth}
         \centering
         \includegraphics[width=\textwidth]{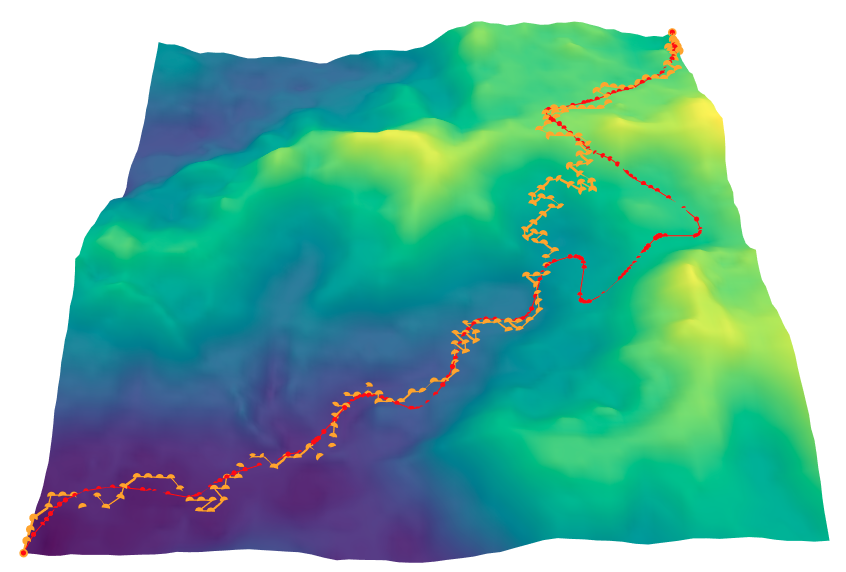}
         \caption{Planned paths for KT-22 height field.}
         \label{fig:kt22_astar}
     \end{subfigure}
     \hfill
     \begin{subfigure}[b]{0.48\textwidth}
         \centering
         \includegraphics[width=\textwidth]{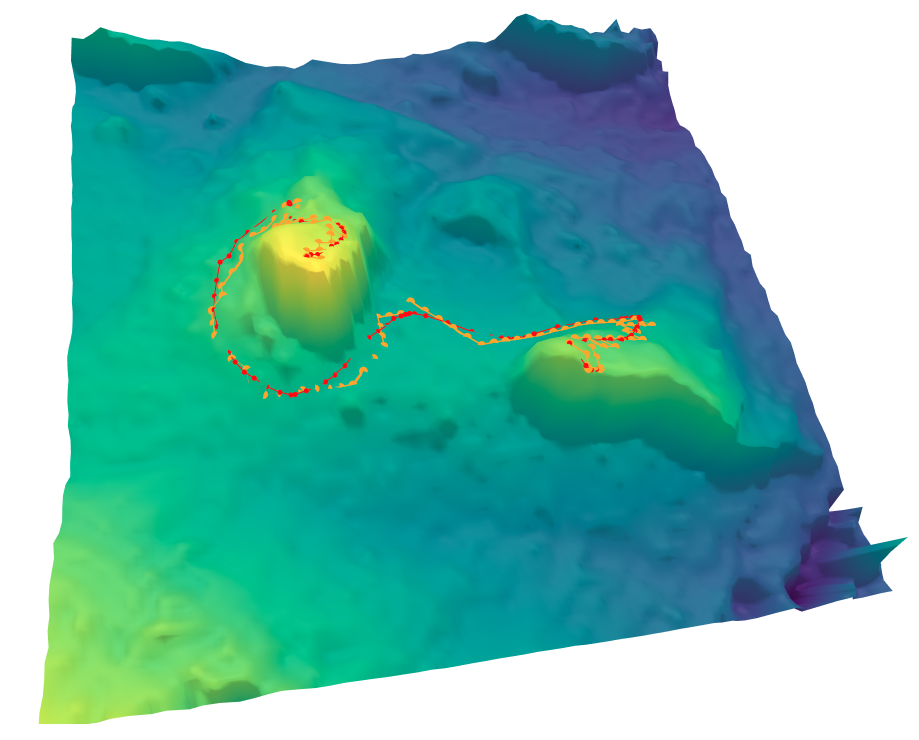}
         \caption{Planned paths for Red Rocks height field.}
         \label{fig:kt22_opt}
     \end{subfigure}
    \caption{Path planning on the NEMo height field. The initial A* path is shown in orange, while the refined path is shown in red. The refined paths from method designed for the implicit NEMo height field are smoother and more efficient. }
    \label{fig:path_planning}
\end{figure}
Fig. \ref{fig:path_planning} shows path planning results for both scenes.
We evaluate the planned paths based on three metrics: 2D path distance, average slope, and smoothness.
Smoothness is computed as estimated jerk from triple differencing the positions (lower is smoother). 
We report a comparison of these metrics in Table~\ref{tab:planning_metrics} for the select paths shown in Fig.~\ref{fig:path_planning}.

The path optimization produces much smoother and dynamically feasible paths. 
The average slope is similar to or slightly higher than that of A*---this is due to A* utilizing switchbacks to minimize vertical transition while sacrificing smoothness, while the path optimization is currently unable to route smooth switchbacks.
This behavior should be possible with further improvement and tuning of the planning algorithm.





\begin{table}[htbp]
\caption{Path planning metrics.}
\begin{subtable}[h]{0.5\textwidth}
    \begin{center}
    \begin{tabular}{|c|c c c|}
        \cline{2-4}
        \multicolumn{1}{c|}{} & Path distance (2D) & Average slope & Smoothness \\
        \hline\hline
        \textbf{A* path}      & 4.892  & 0.347  & 0.0258  \\
        \textbf{Refined path} & 3.278  & 0.365  & 0.0174  \\
        \hline
    \end{tabular}
    \vspace*{1.5mm}
    \caption{KT-22}
    \label{tab:kt22}
    \end{center}
\end{subtable}
\hfill
\begin{subtable}[h]{0.5\textwidth}
    \begin{center}
    \begin{tabular}{|c|c c c|}
        \cline{2-4}
        \multicolumn{1}{c|}{} & Path distance (2D) & Average slope & Smoothness \\
        \hline\hline
        \textbf{A* path}      & 2.932  & 0.398  & 0.0195  \\
        \textbf{Refined path} & 2.172  & 0.412  & 0.0138  \\
        \hline
    \end{tabular}
    \vspace*{1.5mm}
    \caption{Red Rocks}
    \label{tab:redrocks}
    \end{center}
\end{subtable}
\hfill
\label{tab:planning_metrics}
\end{table}

\section{Conclusion}

In this work, we propose Neural Elevation Models, a novel representation for terrain adapted from Neural Radiance Fields.
This representation leverages the advantages of NeRFs in fast generation solely from imagery and rich visual reconstruction, while also capturing compact terrain geometry in the form of a continuous and differentiable height field.
As next steps we plan to distill feature and semantic information from the NeRF into the height network, which can then be readily leveraged for path planning.
In addition, we aim to eventually perform validation of planned paths with rovers in simulated and real-world environments.

\section*{Acknowledgment}
 
The authors would like to thank Professor Mac Schwager for very insightful discussion, and Daniel Neamati for valuable feedback and for reviewing this paper.

\bibliographystyle{IEEEtran}
\bibliography{references}

\begin{thebibliography}{10}
\providecommand{\url}[1]{#1}
\csname url@rmstyle\endcsname
\providecommand{\newblock}{\relax}
\providecommand{\bibinfo}[2]{#2}
\providecommand\BIBentrySTDinterwordspacing{\spaceskip=0pt\relax}
\providecommand\BIBentryALTinterwordstretchfactor{4}
\providecommand\BIBentryALTinterwordspacing{\spaceskip=\fontdimen2\font plus
\BIBentryALTinterwordstretchfactor\fontdimen3\font minus \fontdimen4\font\relax}
\providecommand\BIBforeignlanguage[2]{{%
\expandafter\ifx\csname l@#1\endcsname\relax
\typeout{** WARNING: IEEEtran.bst: No hyphenation pattern has been}%
\typeout{** loaded for the language `#1'. Using the pattern for}%
\typeout{** the default language instead.}%
\else
\language=\csname l@#1\endcsname
\fi
#2}}

\bibitem{ono2016data}
M.~Ono, B.~Rothrock, E.~Almeida, A.~Ansar, R.~Otero, A.~Huertas, and M.~Heverly, ``{Data-Driven Surface Traversability Analysis for Mars 2020 Landing Site Selection},'' in \emph{2016 IEEE Aerospace Conference}.\hskip 1em plus 0.5em minus 0.4em\relax IEEE, 2016, pp. 1--12.

\bibitem{carsten2007global}
J.~Carsten, A.~Rankin, D.~Ferguson, and A.~Stentz, ``{Global Path Planning on Board the Mars Exploration Rovers},'' in \emph{2007 IEEE Aerospace Conference}.\hskip 1em plus 0.5em minus 0.4em\relax IEEE, 2007, pp. 1--11.

\bibitem{schoenberger2016sfm}
J.~L. Sch\"{o}nberger and J.-M. Frahm, ``{Structure-from-Motion Revisited},'' in \emph{Conference on Computer Vision and Pattern Recognition (CVPR)}, 2016.

\bibitem{mildenhall2021nerf}
B.~Mildenhall, P.~P. Srinivasan, M.~Tancik, J.~T. Barron, R.~Ramamoorthi, and R.~Ng, ``{NeRF: Representing Scenes as Neural Radiance Fields for View Synthesis},'' \emph{Communications of the ACM}, vol.~65, no.~1, pp. 99--106, 2021.

\bibitem{gao2022nerf}
K.~Gao, Y.~Gao, H.~He, D.~Lu, L.~Xu, and J.~Li, ``{NeRF: Neural Radiance Field in 3D Vision, A Comprehensive Review},'' \emph{arXiv preprint arXiv:2210.00379}, 2022.

\bibitem{mari2022sat}
R.~Mar{\'\i}, G.~Facciolo, and T.~Ehret, ``{Sat-NeRF: Learning Multi-View Satellite Photogrammetry with Transient Objects and Shadow Modeling using RPC Cameras},'' in \emph{Proceedings of the IEEE/CVF Conference on Computer Vision and Pattern Recognition}, 2022, pp. 1311--1321.

\bibitem{xiangli2022bungeenerf}
Y.~Xiangli, L.~Xu, X.~Pan, N.~Zhao, A.~Rao, C.~Theobalt, B.~Dai, and D.~Lin, ``{BungeeNeRF: Progressive Neural Radiance Field for Extreme Multi-Scale Scene Rendering},'' in \emph{European conference on computer vision}.\hskip 1em plus 0.5em minus 0.4em\relax Springer, 2022, pp. 106--122.

\bibitem{tancik2023nerfstudio}
M.~Tancik, E.~Weber, E.~Ng, R.~Li, B.~Yi, T.~Wang, A.~Kristoffersen, J.~Austin, K.~Salahi, A.~Ahuja, \emph{et~al.}, ``{Nerfstudio: A Modular Framework for Neural Radiance Field Development},'' in \emph{ACM SIGGRAPH 2023 Conference Proceedings}, 2023, pp. 1--12.

\bibitem{koenker2001quantile}
R.~Koenker and K.~F. Hallock, ``{Quantile Regression},'' \emph{Journal of economic perspectives}, vol.~15, no.~4, pp. 143--156, 2001.

\bibitem{yang2016posterior}
Y.~Yang, H.~J. Wang, and X.~He, ``{Posterior Inference in Bayesian Quantile Regression with Asymmetric Laplace Likelihood},'' \emph{International Statistical Review}, vol.~84, no.~3, pp. 327--344, 2016.

\bibitem{adamkiewicz2022vision}
M.~Adamkiewicz, T.~Chen, A.~Caccavale, R.~Gardner, P.~Culbertson, J.~Bohg, and M.~Schwager, ``{Vision-Only Robot Navigation in a Neural Radiance World},'' \emph{IEEE Robotics and Automation Letters}, vol.~7, no.~2, pp. 4606--4613, 2022.

\bibitem{levine2011necessary}
J.~L{\'e}vine, ``{On Necessary and Sufficient Conditions for Differential Flatness},'' \emph{Applicable Algebra in Engineering, Communication and Computing}, vol.~22, no.~1, pp. 47--90, 2011.

\bibitem{GES}
\BIBentryALTinterwordspacing
``{Google Earth Studio}.'' [Online]. Available: \url{https://earth.google.com/studio/}
\BIBentrySTDinterwordspacing

\bibitem{dronemapper}
``{DroneMapper Sample Data},'' \url{https://dronemapper.com/sample_data/}, accessed: March 19, 2024.

\bibitem{muller2022instant}
T.~M{\"u}ller, A.~Evans, C.~Schied, and A.~Keller, ``{Instant Neural Graphics Primitives with a Multiresolution Hash Encoding},'' \emph{ACM transactions on graphics (TOG)}, vol.~41, no.~4, pp. 1--15, 2022.

\bibitem{kingma2014adam}
D.~P. Kingma and J.~Ba, ``{Adam: A Method for Stochastic Optimization},'' \emph{arXiv preprint arXiv:1412.6980}, 2014.

\end{thebibliography}

\end{document}